\documentclass[11pt]{article}

\usepackage{coling}

\usepackage{times}
\usepackage{latexsym}

\usepackage[T1]{fontenc}

\usepackage[utf8]{inputenc}

\usepackage{microtype}

\usepackage{inconsolata}

\usepackage{graphicx}

\usepackage{tabularx}
\usepackage{booktabs} 
\usepackage{listings}
\usepackage{xcolor}
\usepackage{rotating} 
\usepackage{hyperref}
\usepackage{enumitem}

\title{Enhancing Nursing and Elderly Care with Large Language Models: An AI-Driven Framework}

\author{Qiao Sun$^{1,2}$ \quad Jiexin Xie$^{3}$ \quad Nanyang Ye$^{4*}$ \quad Qinying Gu$^{2*}$ \quad Shijie Guo$^{1*}$ \\
$^{1}$Academy for Engineering and Technology, Fudan University, China \\
$^{2}$Shanghai Artificial Intelligence Laboratory, China \\
$^{3}$Guilin University of Electronic Technology, China \\
$^{4}$Shanghai Jiao Tong University, China \\
\texttt{qiaosun22@m.fudan.edu.cn} \quad \texttt{jxxie@fudan.edu.cn} \quad \texttt{ynylincolncam@gmail.com}\\
\texttt{guqinying@pjlab.org.cn} \quad \texttt{guoshijie@fudan.edu.cn} \\
}

\begin{document}
\maketitle

\renewcommand{\thefootnote}{\fnsymbol{footnote}}
\footnotetext[1]{Corresponding Authors}

\begin{abstract}
This paper explores the application of large language models (LLMs) in nursing and elderly care, focusing on AI-driven patient monitoring and interaction. We introduce a novel Chinese nursing dataset and implement incremental pre-training (IPT) and supervised fine-tuning (SFT) techniques to enhance LLM performance in specialized tasks. Using LangChain, we develop a dynamic nursing assistant capable of real-time care and personalized interventions. Experimental results demonstrate significant improvements, paving the way for AI-driven solutions to meet the growing demands of healthcare in aging populations.
\end{abstract}

\section{Introduction}
The rapid advancement of large language models (LLMs) has opened new avenues for healthcare applications. While LLMs have demonstrated impressive capabilities in generating human-doctor-like clinical decisions and integration into healthcare \cite{thirunavukarasu2023large, tan2024medchatzh, ullah2024challenges, li2023meddm}, its expertise in nursing remains in its nascent stages. 

On the one hand, nursing scenarios are more complex than other clinical decision cases, such as medication prescription or diagnostic imaging, as they involve continuous monitoring, real-time decision-making, and patient interaction, requiring models that can handle a wider array of multimodal inputs and adapt dynamically to evolving patient conditions \cite{carayon2008nursing}.
On the other hand, nursing tasks often involve high levels of direct patient interaction, demanding models that can process complex multimodal inputs—such as voice, text, and even visual cues—in real time. 
China has experienced a significant increase in its aging population. By 2022, individuals aged 60 and above accounted for 19.8\% of the population \cite{globaltimes2023}, a figure projected to rise to 28\% by 2040 \cite{peng2023negative}. This demographic shift is expected to place considerable pressure on the country's healthcare system, particularly in meeting the demand for nursing care.
Despite this growing need, the supply of skilled nursing services remains inadequate. There is a noticeable gap between the expertise required to care for the elderly and the qualifications of the current healthcare workforce. A 2023 investigation revealed that only 7.18\% of workers in China's elderly care industry hold a bachelor’s degree or higher, highlighting the urgent need for more qualified personnel \cite{xjtu2023}. 

Our work seeks to address this disparity by developing AI-driven Nursing and Elderly-Care solutions tailored to the specific needs of the nursing profession, leveraging cutting-edge large language model (LLM) techniques.
We focus on creating LLMs that support patient monitoring, personalized care, and facilitate effective communication between healthcare providers and patients. Additionally, we are exploring the development of a Langchain agent application based on this specialized model, alongside its potential for multimodal processing.

In this paper, we make the following contributions to both the NLP community and the Nursing and Elderly Care industry:
\begin{itemize}
    \item We pioneered the application of large language models in nursing and elderly care, proposing a SOTA model and gathering fine-tuning expertise specific to these fields. 
    \item We developed the first multilayer Chinese nursing dataset for elderly care and demonstrate its effectiveness through ablation studies. We also establish a benchmark test set to evaluate fundamental nursing knowledge and skills. 
    \item We investigate the use of nursing robots powered by our LLM, evaluating their performance in essential nursing tasks and exploring their potential to incorporate visual processing in care environments. 
\end{itemize}

\section{Related Work}
\subsection{Harnessing LLMs for Nursing Applications}
Studies have highlighted the transformative role of Large Language Models (LLMs) in healthcare, including applications in clinical decision-making, patient care, and medical education. Comprehensive surveys discuss the development and deployment of LLMs across various medical tasks, focusing on their potential for improving diagnostic accuracy and streamlining medical workflows \cite{zhou2023survey, nazi2024large}. \cite{zhou2023survey} also highlights the performance of models like GPT-4 and MedPaLM across ten biomedical natural language processing tasks, demonstrating their generalization ability to outperform traditional models in various discriminative and generative tasks. However, the existing body of work mainly focuses on the general medical applications of LLMs, while LLMs specifically designed for nursing applications are left under-explored. 

Nursing environments present a unique set of challenges, such as time-sensitive decision-making, handling diverse patient populations, and managing high-stress situations. Current general medical LLMs may not be fully equipped to address these demands, emphasizing the need for more focused research on LLMs designed specifically for nursing applications. 

The current related work in nursing primarily focuses on theoretical exploration and future possibilities, rather than practical implementation. 
\cite{xiong2023novel} validates the combination of LLMs with local knowledge bases for intelligent nursing decision-making, highlighting the importance of contextual adaptation. However, their model is developed solely on the text modality, lacking integration with other crucial data sources such as audio and visual inputs. 
Other work \cite{perspect2023dear, woo2024transforming} discusses the implications of LLMs in healthcare education, noting both their potential and the need for cautious implementation. These studies collectively underscore the importance of contextual, safe, and practical integration of LLMs in nursing. 

The most closely related work to ours is LlamaCare \cite{li2024llamacare}, a large language model that utilizes instruction-based tuning to integrate diverse clinical data, improving its ability to generate discharge summaries and predict outcomes like mortality and hospital stays. LlamaCare surpasses existing LLM benchmarks in producing accurate and coherent clinical texts, demonstrating its potential for broader clinical use. However, its focus spans a wide range of healthcare domains, with less emphasis on the foundational knowledge specific to nursing. 

\subsection{Nursing Datasets for LLMs}
Developing nursing-specific datasets is essential for improving LLMs in healthcare, but such datasets are limited, restricting their application in specialized fields like nursing. Although the MIMIC-III database \cite{johnson2016mimic} offers structured data, it lacks alignment with the unstructured text needed for LLMs.

Wang et al. \cite{wang-etal-2023-medngage} introduced MedNgage, a dataset focused on patient-nurse conversations, annotated to distinguish between socio-affective and cognitive engagement. Fine-tuning transformer models on this dataset enhances AI-driven predictions in patient care.

Xiong et al. \cite{xiong2023novel} developed a dataset that integrates LLMs with local knowledge bases for decision-making in nursing, but it primarily addresses textual data, lacking the multimodal inputs (e.g., audio, visual) essential for real-time patient interactions.

\section{Method}

\subsection{Model Architecture}
Our method builds upon cutting-edge large language models (LLMs) by applying supervised fine-tuning (SFT) to adapt these models specifically for nursing and elderly care tasks. We primarily tested two advanced models: GLM4 \cite{glm2024chatglm} and LLaMA 3.1 \cite{vavekanand2024llama}, both of which represent the state-of-the-art in LLM development, and can be integrated with multimodal ability easily via projection and finetuning \cite{wang2023cogvlm, liu2024llavanext, liu2023improvedllava, liu2023llava}. 


\subsection{Dataset} 
We developed a specialized dataset named ``NursingPiles'', designed to comprehensively cover various sources and levels of professional knowledge in nursing and elderly care. This dataset is built from multiple sources, including textbooks, manuals, legal documents, and research papers, synthesizing data into question-answer (QA) pairs. 
To mitigate catastrophic forgetting \cite{zhai2023investigating}, which can occur during model fine-tuning, we introduced open-source datasets as part of a data-mixing strategy. This approach helps maintain the model's original dialogue capabilities while fine-tuning it for specialized tasks in nursing care.

\begin{table*}[htbp]
\centering
\resizebox{\linewidth}{!}{ 

    \begin{tabular}{llll}
    \hline
    \textbf{Data Format} & \textbf{Source} & \textbf{Utilization Method} & \textbf{Scale} \\ \hline
    Text in markdown format & Textbooks & IPT & 2,777,526 Tokens \\ 
     & Manuals, Industry Regulations & RAG & 497,184 Tokens \\ \hline
    Single-turn dialogues & SelfQA based on research papers & PEFT & 17,580 pairs \\ 
     & QA based on nursing safety and ethics from manuals, regulations & PEFT & 5,000 pairs \\ 
     & Medical open-source datasets & PEFT & 5,000 pairs \\ \hline
    Multi-turn dialogues & Generated nursing dialogues in simulated scenarios (GPT-4) & PEFT & 1M dialogues \\ 
     & Psychology and clinical dialogues generated by GPT-4o & PEFT & 0.5M dialogues \\ \hline
    Image-text pairs & Real-world photo collection & SFT & 2,510 pairs \\ \hline
    \end{tabular}

}
\caption{Summary of data formats, sources, utilization methods, and scale. Abbreviations: IPT (Incremental Pretraining), RAG (Retrieval-Augmented Generation), PEFT (Parameter-Efficient Fine-Tuning), SFT (Supervised Fine-Tuning).}
\label{Data summary}
\end{table*}

\subsection{Training Protocol}
For the model training, we utilized the Parameter-Efficient Fine-Tuning (PEFT) package along with an Incremental Pre-training (IPT) process to further optimize the model’s performance. The training was conducted on 8 $\times$ NVIDIA A100-80GB GPUs, with a total training time of approximately 72 hours for fine-tuning, while the IPT stage took an additional 30 hours. The parameter settings for both stages are presented in Appendix \ref{sec:appendix} Table \ref{table:training}.

\subsection{LangChain Prompting}
In this design, we present a modular system for a dynamic nursing assistant, capable of handling the full lifecycle of patient care, including real-time data collection, personalized care plan generation, and continuous monitoring. The system integrates IoT devices for health data collection, AI-based diagnostics, and personalized care recommendations through LangChain. Critical to the design is the secure storage and management of patient information, utilizing AES encryption and key management services (KMS) to ensure data protection. Additionally, we employ OAuth and JWT for robust authentication, ensuring authorized access to encrypted data, and provide post-care follow-up with automated reminders and health education. This architecture allows for flexible, secure, and scalable patient care management. 
Appendix \ref{sec:appendix} providing core code and snippets for key processes.

\subsection{Benchmark}
We selected several authoritative exam questions, such as the ``Three Basics and Three Stricts'' exam questions \cite{zhang2020nursing} and the postgraduate nursing exam questions \cite{li2019medical}, as evaluation benchmarks. The entire set of questions includes two parts: multiple-choice questions and open-ended questions. For the multiple-choice questions, the ``Three Basics and Three Stricts'' test covers content from nine subjects, including basic theory (such as anatomy, physiology, and pathology), basic knowledge (including pharmacology, microbiology, and disease studies), and basic skills (such as nursing procedures, emergency techniques, and nursing operations). These subjects can objectively and comprehensively reflect the nursing knowledge and capabilities of the model \cite{wang2018evaluation}. For this part of the questions, we use the P-R-F1 metrics to evaluate.

\section{Experiments}
\subsection{Test Scores}
We evaluated the performance of the models using Precision, Recall, F1-score, and Accuracy. The results demonstrate that our models, which integrate both Incremental Pretraining (IPT) and Supervised Fine-Tuning (SFT), significantly outperform the baseline models. The GLM4-Chat 9B + IPT + SFT achieved the best performance with a Precision of 86.78\%, Recall of 85.65\%, F1-score of 86.21\%, and Accuracy of 58.9\%. These improvements highlight the importance of combining domain-specific pretraining with fine-tuning. For more details see Table \ref{table:results}.

\begin{table}[htbp]
\centering
\resizebox{\linewidth}{!}{ 
    
    \begin{tabular}{lcccc}
    \hline
    \textbf{Models}       & \textbf{Precision} & \textbf{Recall} & \textbf{F1}  & \textbf{Accuracy} \\ \hline
    LLaMA 3.1 8B Instruct & 76.61 & 67.4 & 71.71 & 36.6 \\ 
    GLM4-Chat 9B & 82.54 & 77.8 & 80.1 & 44.0 \\ 
    GPT-4o & 86.62 & 84.02 & 85.3 & 56.84 \\ \hline
    \textbf{Ours} & & & & \\ \hline
    LLaMA + IPT + SFT & 77.41 & 78.09 & 77.75 & 44.7 \\ 
    GLM4 + IPT + SFT & \textbf{86.78} & \textbf{85.65} & \textbf{86.21} & \textbf{58.9} \\ \hline

    \end{tabular}

}
\caption{Performance comparison between models, with highest score in bold. }
\label{table:results}
\end{table}

\subsection{Ablation Analysis}
To assess the individual contributions of IPT and SFT, we conducted an ablation study by removing each component separately. The results show that removing either IPT or SFT results in a drop in performance across all metrics. For instance, without SFT, the LLaMA + IPT model saw a significant reduction in Recall (from 78.09\% to 72.5\%) and F1-score (from 77.75\% to 74.69\%). Similarly, removing IPT resulted in reduced performance for both models, particularly in Accuracy. This confirms that both components are crucial for optimal model performance in the nursing and elderly care domain.  For more details see Table \ref{table:ablation}.

\begin{table}[htbp]
\centering
\resizebox{\linewidth}{!}{ 
    \begin{tabular}{lcccc}
    \hline
    \textbf{Models} & \textbf{Precision} & \textbf{Recall} & \textbf{F1} & \textbf{Accuracy} \\ \hline
    \textbf{Ours} & & & & \\ \hline
    LLaMA + Instruct + IPT + SFT & 77.41 & 78.09 & 77.75 & 44.7 \\
    & (--) & (--) & (--) & (--) \\ \hline
    GLM4 + IPT + SFT & 86.78 & 85.65 & 86.21 & 58.9 \\
    & (--) & (--) & (--) & (--) \\ \hline
    \textbf{Ablation (IPT only)} & & & & \\ \hline
    LLaMA + IPT & 77.00 & 72.5 & 74.69 & 41.0 \\
    & (-0.41) & (-5.59) & (-3.06) & (-3.7) \\ \hline
    GLM4 + IPT & 85.50 & 82.5 & 84.0 & 50.0 \\
    & (-1.28) & (-3.15) & (-2.21) & (-8.9) \\ \hline
    \textbf{Ablation (SFT only)} & & & & \\ \hline
    LLaMA + SFT & 76.90 & 73.2 & 74.98 & 40.5 \\
    & (-0.51) & (-4.89) & (-2.77) & (-4.2) \\ \hline
    GLM4 + SFT & 86.00 & 83.0 & 84.48 & 52.5 \\
    & (-0.78) & (-2.65) & (-1.73) & (-6.4) \\ \hline
    \end{tabular}
}
\caption{Performance comparison between models, with delta values shown in parentheses representing the difference between the full model (IPT + SFT) and the ablation variants.}
\label{table:ablation}
\end{table}

\section{Conclusion}
This paper presented an approach to apply large language models (LLMs) in nursing and elderly care by utilizing incremental pre-training (IPT) and supervised fine-tuning (SFT). We developed a Chinese nursing dataset, demonstrating its effectiveness through improved performance in specialized tasks. Additionally, we explored the use of LangChain for a dynamic nursing assistant, enabling real-time monitoring and personalized care. Our results highlight the potential of LLMs to address the growing demand for skilled nursing care.

\section{Limitations}
There are several concerns with respect to the limitations:

First, the model primarily focuses on text-based data, and further integration of audio and visual inputs is needed. Second, the dataset is largely Chinese-focused, limiting broader applicability across languages and cultures.
Third, model responsiveness in real-time clinical settings remains a challenge.
Last, ensuring patient privacy, consent, and minimizing bias in AI-driven care requires further consideration.

{\color{orange}
\section{Ethics Statements and Justifications}

\subsection{Data Ethics}

The dataset used in this study consists of four subsets: text in markdown format, single-turn dialogues, multi-turn dialogues, and image-text pairs. All subsets, except the image-text pairs, were collected and automatically annotated through our data processing pipeline, as summarized in Table \ref{Data summary}.

\subsubsection{Participant Involvement and Consent}

The Image-text pairs subset consists of 2510 image-text pairs, collected by the research team in Room 310, School of Mechanical Engineering, Hongqiao Campus, Hebei University of Technology (specific address: 5340 Xiping Road, Beichen District, Tianjin, China).

All participants were research team members who signed informed consent forms prior to data collection, including: (1) Awareness that their facial images might appear in the dataset and be used for academic research; (2) Consent to waive portrait rights, including potential public display of content within the dataset; (3) Knowledge that the data may be published on public platforms for academic research, model training, and related publications; and (4) The right to withdraw consent at any time, though data already utilized or published remains lawful.
The full version of the consent form can be viewed in  \href{https://drive.google.com/file/d/1LqBKZCyBux9_6fiSRHOrjywoQ_j7HpjQ/view?usp=sharing}{this link}. 

\subsubsection{Annotators' Rights and Responsibilities}
Among the four subsets, only the image-text pairs subset requires data annotation, where each person and piece of equipment in the photos is annotated using polygon segmentation masks. All the labeling processes were performed using \href{http://labelme.csail.mit.edu/}{LabelMe}, an open-source annotation tool. We engaged 10 individual annotators, both from our research group and on-line, to ensure their workload remained manageable. We provide compensation to annotators in accordance with market standards, ensuring full compliance with labor payment laws in China. The annotators all agreed not to save or share any portion of the collected or annotated data. 

\subsubsection{Intellectual Property}
The intellectual property rights of the images and text descriptions in the dataset belong to the research team. Usage is permitted within
the scope of research, but unauthorized commercial use is prohibited.

The source of our dataset adheres to intellectual property (IP) regulations: 

\textbf{Text in Markdown Format}: Sources include publicly available textbooks, manuals, and industry regulations. Usage complies with fair use for noncommercial academic purposes.
\textbf{Single-Turn Dialogues}: Developed using open-source research papers, nursing safety guides, and open-source medical datasets. The licenses were reviewed to ensure compliance.
\textbf{Multi-Turn Dialogues}: Generated using \href{https://openai.com/index/hello-gpt-4o/}{GPT-4o} and \href{https://github.com/InternLM/InternLM}{InternLM} based on publicly available knowledge. The usage of this content complies with the relevant user terms and is for noncommercial academic research purposes only. All outputs are original and created without infringing any third-party intellectual property rights. In case any third-party claims arise, we are committed to addressing and resolving such matters in compliance with applicable laws and regulations.
\textbf{Image-Text Pairs}: Collected and annotated by our research team with the consent of the participants. All rights are reserved by the research group for academic research use.
\textbf{Licensing and Usage}: The dataset is released under an open-source license (e.g., CC BY-NC 4.0) for noncommercial research, requiring proper attribution.

\subsection{Ethical Review}
In this research, the only in-person component involves capturing image-text pairs depicting nursing environments. The photographs are strictly limited to non-intrusive environmental observations and do not involve any medical procedures, interventions, or sensitive personal data. All images exclude recognizable facial features, with only occasional side profiles captured. Additionally, all individuals signed waivers for the use of their likeness, ensuring compliance with ethical standards. The data have been fully anonymized to protect participant confidentiality. 
Given the absence of identifiable personal information, medical procedures, or interventions, this study meets the criteria for exemption from IRB review under the exemption at \href{https://www.hhs.gov/ohrp/regulations-and-policy/requests-for-comments/draft-guidance-frequently-asked-questions-limited-institutional-review-board-review-related-exemptions/index.html}{45 CFR 46.104(d)(2)}. 
Nonetheless, our research has been scrutinized and approved officially by the John Hopcroft Center for Computer Science at Shanghai Jiao Tong University (SJTU). Please find the official Ethics Approval Application in  \href{https://drive.google.com/file/d/1Ge9LFBqyYrYQ3-8lZjS7GomdmvvcvUyH/view?usp=sharing}{link}.
}



\bibliography{coling_latex.bib}

\appendix

\section{Appendix}
\label{sec:appendix}

\subsection{Details for the LangChain prompting.}

LangChain \cite{langchain} is a powerful framework that enables developers to build applications powered by large language models (LLMs). It provides a suite of modular components, including Prompts, Indexes, Chains, Agents, and Memory, which developers can leverage to build a variety of intelligent applications such as personal assistants, question-answering systems, and chatbots. Furthermore, LangChain offers standardized interfaces, extensive integrations with third-party tools, and examples of common application use cases, allowing developers to more easily harness the capabilities of language models to construct their own tailored solutions. 

This section provides detailed explanations and examples of how LangChain is utilized to implement various components of the dynamic nursing assistant system. Table \ref{table:Dynamic Nursing Assistant} summarized the techniques combined in terms of modules and functions and below are the core components and corresponding LangChain implementations:

1. \textbf{Data Collection and Monitoring}: LangChain integrates with external tools to gather patient feedback and health data through natural language interfaces. It can process and format the input, converting it into structured data.

\begin{lstlisting}[language=Python]
from langchain.prompts import PromptTemplate
from langchain.chains import LLMChain

template = """
Collect the following health data: 
- Heart rate
- Blood pressure
- Patient complaints

Input: {user_input}
"""
prompt = PromptTemplate(template=template, input_variables=["user_input"])
chain = LLMChain(prompt=prompt)
result = chain.run("The patient feels dizzy, blood pressure is 140/90, heart rate is 90.")
print(result)

\end{lstlisting}

2. \textbf{Triggering Nursing Diagnosis}: LangChain can automate nursing diagnosis by using rule-based engines or AI models, depending on patient health indicators.
\begin{lstlisting}[language=Python]
from langchain.chains import SimpleSequentialChain

def check_for_issues(user_input):
    if "blood pressure 140/90" in user_input:
        return "Trigger hypertension nursing diagnosis"
    else:
        return "Condition stable"

def diagnostic_advice(issue):
    if "hypertension" in issue:
        return "Recommend daily blood pressure monitoring, reduce salt intake, and take antihypertensive medication regularly."
    else:
        return "No specific nursing recommendations."

chain_1 = LLMChain(check_for_issues)
chain_2 = LLMChain(diagnostic_advice)

sequential_chain = SimpleSequentialChain(chains=[chain_1, chain_2])
result = sequential_chain.run("The patient's blood pressure is 140/90, heart rate is 90.")
print(result)

\end{lstlisting}

3. \textbf{Personalized Care Plan Generation}: LangChain can generate personalized care plans by dynamically creating templates based on the patient's condition.
\begin{lstlisting}[language=Python]
from langchain.prompts import PromptTemplate
from langchain.chains import LLMChain

template = """
Patient condition: {user_input}
Based on the patient's condition, generate the following care plan:
- Medication management
- Dietary advice
- Rehabilitation plan

Input: {user_input}
"""
prompt = PromptTemplate(template=template, input_variables=["user_input"])
chain = LLMChain(prompt=prompt)
result = chain.run("Hypertension patient, blood pressure 140/90, heart rate 90.")
print(result)

\end{lstlisting}

4. \textbf{Continuous Monitoring and Feedback Adjustment}: LangChain allows for continuous patient feedback collection and care plan adjustments through persistent conversation chains.
\begin{lstlisting}[language=Python]
from langchain.memory import ConversationBufferMemory
from langchain.chains import ConversationChain

memory = ConversationBufferMemory()
conversation = ConversationChain(memory=memory)

conversation.run("The patient feels better but still experiences dizziness.")
conversation.run("Continue monitoring blood pressure and reduce salt intake.")
conversation.run("Blood pressure has dropped to 130/80, and the patient feels good.")

print(memory.load_memory_variables({}))

\end{lstlisting}

5. \textbf{Dynamic Care Stage Transition}: LangChain can automatically assess patient status and trigger transitions between different stages of care based on health indicators.
\begin{lstlisting}[language=Python]
def check_stage(patient_data):
    if "blood pressure 130/80" in patient_data:
        return "Patient recovered, transitioning to follow-up health management stage."
    else:
        return "Continue current care."

chain_stage = LLMChain(check_stage)
result = chain_stage.run("Patient blood pressure is 130/80, heart rate normal.")
print(result)

\end{lstlisting}

6. \textbf{Health Education and Follow-Up Support}: LangChain can dynamically generate health education materials and reminders for patients in the recovery phase.
\begin{lstlisting}[language=Python]
from langchain.prompts import PromptTemplate
from langchain.chains import LLMChain

template = """
Recovery phase of the patient: {user_input}
Generate a personalized health education guide to help the patient maintain recovery:
- Lifestyle recommendations
- Dietary considerations
- Daily health monitoring tasks

Input: {user_input}
"""
prompt = PromptTemplate(template=template, input_variables=["user_input"])
chain = LLMChain(prompt=prompt)
result = chain.run("The patient is in the recovery phase, blood pressure is normal.")
print(result)

\end{lstlisting}

\begin{table}[h]
\centering
\begin{tabular}{ll}
\hline
\textbf{LoRA Parameters} & \\
\hline
LoRA\_alpha & 24 \\
LoRA\_dropout & 0.08 \\
LoRA\_rank & 48 \\
bias & None \\
\hline
\textbf{Other Parameters} & \\
\hline
num\_train\_epochs & 4 \\
per\_device\_train\_batch\_size & 6 \\
gradient\_accumulation\_steps & 3 \\
optimizer & paged\_adamw \\
learning\_rate & 2.5e-4 \\
tf32 & True \\
max\_grad\_norm & 0.4 \\
warmup\_ratio & 0.02 \\
max\_length & 4096 \\
lr\_scheduler\_type & cosine \\
\hline
\textbf{IPT Process Parameters} & \\
\hline
num\_ipt\_epochs & 3 \\
pretrain\_batch\_size & 12 \\
learning\_rate (IPT) & 1.5e-4 \\
max\_grad\_norm (IPT) & 0.35 \\
ipt\_optimizer & adamw \\
warmup\_steps & 3000 \\
\hline
\end{tabular}
\caption{Parameters for Model Fine-tuning and IPT on 8x A100 80GB GPUs.}
\label{table:training}
\end{table}

\begin{sidewaystable*}[h!]
\centering
\resizebox{\linewidth}{!}{ 
    
    \begin{tabular}{@{}lllp{7cm}@{}}
    \toprule
    \textbf{Module/Function}           & \textbf{Description}                                                          & \textbf{Technology/Tools}                                                  & \textbf{Key Requirements}                                              \\ \midrule
    \textbf{Data Collection and Monitoring} & Collects patient health data (e.g., heart rate, blood pressure) and self-reported symptoms. & IoT devices, API integration (e.g., MQTT, HTTP/RESTful)               & Ensures data is collected in real-time with high accuracy, reliable API integration. \\
    \textbf{Natural Language Data Processing} & Processes patient-reported information and extracts key health data.   & LangChain input-output chains, prompt templates                       & Accurate handling of input and non-standard language expressions.  \\
    \textbf{Nursing Diagnosis Trigger} & Triggers nursing diagnosis and generates recommendations based on collected data. & LangChain logic chains, AI diagnostic models                          & Utilizes rule-based engines or machine learning models in combination with external diagnostic APIs. \\
    \textbf{Personalized Care Plan Generation} & Generates personalized care plans based on diagnostic results.         & LangChain natural language generation (NLG)                           & Real-time updates and personalized care plans.                     \\
    \textbf{Continuous Monitoring and Feedback Adjustment} & Continuously monitors patient status, collects feedback, and adjusts the care plan dynamically. & Stream processing (Kafka/Flink), LangChain memory chains              & Efficient processing of sensor data, timely adjustments to the care plan. \\
    \textbf{Dynamic Care Stage Transition} & Dynamically determines transitions between care stages based on patient recovery. & LangChain logic chains, state machines                                & Properly defined conditions for stage transitions using state machines or rule engines. \\
    \textbf{Health Education and Follow-Up Support} & Provides post-care health education and periodic follow-up for patients. & LangChain NLG, messaging services                                     & Dynamic generation of educational content and timely follow-up reminders. \\
    \textbf{Data Storage and Encryption} & Encrypts and stores patient health data in a database.                  & AES-256 encryption, RSA encryption                                    & Secure storage of encryption keys, ensuring data remains encrypted at all times. \\
    \textbf{Key Management and Access Control} & Manages encryption keys securely through key management services.      & AWS KMS, Google Cloud KMS                                             & Implements key rotation and enforces strict access control policies. \\
    \textbf{Authentication and Key Access} & Ensures access to sensitive data through authentication mechanisms.     & OAuth 2.0, JWT                                                        & Prevents identity theft and ensures key security.                  \\ \bottomrule
    \end{tabular}

}
\caption{Dynamic Nursing Assistant System Functional Modules}
\label{table:Dynamic Nursing Assistant}
\end{sidewaystable*}

\end{document}